\pdfoutput=1
\documentclass[11pt]{article}

\usepackage[final]{acl}
\usepackage[normalem]{ulem} 

\usepackage{times}
\usepackage{latexsym}
\usepackage{placeins}
\usepackage{subcaption}
\usepackage{stfloats}
\usepackage{amsmath}
\usepackage{amssymb}
\usepackage{bbm} 
\usepackage{forest}
\usetikzlibrary{shadows, arrows.meta}
\usepackage{tikz}
\usetikzlibrary{shapes.geometric, positioning, calc}
\usepackage[T1]{fontenc}

\usepackage[utf8]{inputenc}
\usepackage{tabularx} 
\usepackage{microtype}

\usepackage{inconsolata}

\usepackage{graphicx}
\usepackage{booktabs}
\usepackage{multirow}
\usepackage{makecell}
\usepackage{pifont}
\newcommand{\cmark}{\ding{51}}
\newcommand{\xmark}{\ding{55}}
\newcommand{\ours}{\textsc{ours}}
\newcommand{\best}[1]{\textbf{#1}}
\newcommand{\second}[1]{\uline{#1}}
\usepackage{multirow,hhline,makecell,xcolor,colortbl}
\newcommand{\darkblue}[1]{\textcolor{blue!60!black}{#1}}

\title{INFACT: A Diagnostic Benchmark for Induced Faithfulness and Factuality Hallucinations in Video-LLMs}

\author{
 \textbf{Junqi Yang\textsuperscript{1,2}},
 \textbf{Yuecong Min\textsuperscript{1}},
 \textbf{Jie Zhang\textsuperscript{1}},
 \textbf{Shiguang Shan\textsuperscript{1}},
 \textbf{Xilin Chen\textsuperscript{1}},
\\
\\
 \textsuperscript{1}State Key Laboratory of AI Safety, Institute of Computing Technology, Chinese Academy of Sciences
    \\
 \textsuperscript{2}School of Advanced Interdisciplinary Sciences, UCAS
 \\
\\
}

\begin{document}
\maketitle
\begin{abstract}
Despite rapid progress, Video Large Language Models (Video-LLMs) remain unreliable due to hallucinations, which are outputs that contradict either video evidence (faithfulness) or verifiable world knowledge (factuality).
Existing benchmarks provide limited coverage of factuality hallucinations and predominantly evaluate models only in clean settings.
We introduce \textsc{INFACT}, a diagnostic benchmark comprising 9{,}800 QA instances with fine-grained taxonomies for faithfulness and factuality, spanning real and synthetic videos.
\textsc{INFACT} evaluates models in four modes: Base (clean), Visual Degradation, Evidence Corruption, and Temporal Intervention for order-sensitive items.
Reliability under induced modes is quantified using Resist Rate (RR) and Temporal Sensitivity Score (TSS).
Experiments on 14 representative Video-LLMs reveal that higher Base-mode accuracy does not reliably translate to higher reliability in the induced modes, with evidence corruption reducing stability and temporal intervention yielding the largest degradation.
Notably, many open-source baselines exhibit near-zero TSS on factuality, indicating pronounced temporal inertia on order-sensitive questions.

\end{abstract}

\begin{figure*}
    \centering
  \includegraphics[
    width=\linewidth,
    trim=5mm 20mm 10mm 20mm,
    clip
  ]{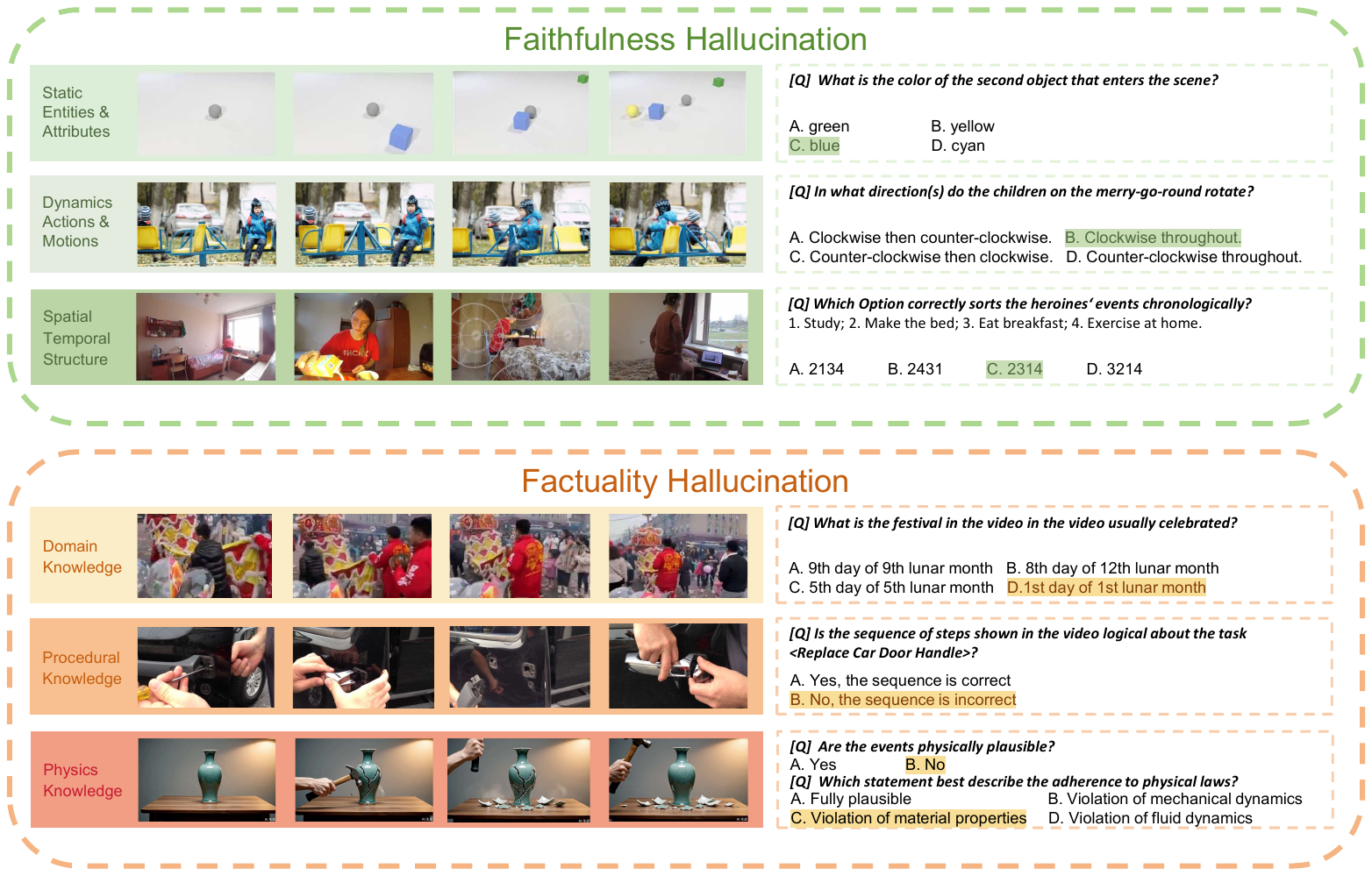}
\caption{\textbf{Examples of Faithfulness and Factuality Hallucinations.} \textbf{Top: Faithfulness} items are verified by video evidence, covering Static Entities \& Attributes, Dynamic Actions \& Motions, and Spatio-Temporal Relations.
\textbf{Bottom: Factuality} items require consistency with world knowledge and cover Domain Knowledge (Know-WHAT), Procedural Knowledge (Know-HOW), and Physical Knowledge (Know-WHY).}
    \label{fig:Factuality_examples}
\end{figure*}

\section{Introduction}

Video Large Language Models (Video-LLMs)~\cite{openai_gpt51_2025, gemini3flash, Qwen3-VL,wang2025internvl35} have made rapid progress in video understanding in recent years, demonstrating impressive capabilities across broad tasks~\cite{mvbench2024, videomme2025_CVPR, ICLR2025_tomato, shafique2025culturallydiversemultilingualmultimodalvideo, eccv_cityguessr}.
Despite these advances, the reliability of Video-LLMs in downstream applications is compromised by hallucinations~\cite{vidhalluc, zhang2024eventhallusion}, generating content that contradicts the provided video evidence (faithfulness) or verifiable world knowledge (factuality).

Recent benchmarks have begun to probe hallucinations in Video-LLMs, e.g., by targeting event/motion-centric failures~\cite{zhang2024eventhallusion,Kong_Zeng_Chen_Li_Yan_Zhu_2025_mhbench}, using controlled contrastive setups~\cite{vidhalluc}, or studying prior-driven shortcuts~\cite{Bae_2025_CVP_MASH_VLM}.
However, existing efforts predominantly emphasize video-verifiable inconsistencies, leaving factuality hallucinations substantially under-explored.
Moreover, high performance in clean scenarios does not guarantee low hallucination rates, as models may exploit shortcuts such as language priors or static cues.
This motivates the evaluation of hallucinations beyond clean settings through controlled evidence perturbations.

To bridge these gaps, we introduce \textsc{INFACT}, a diagnostic benchmark for evaluating Video-LLMs hallucinations regarding faithfulness and factuality in both clean and noisy scenarios. Specifically, \textsc{INFACT} establishes fine-grained taxonomies and comprises 9{,}800 QA instances sapnning real and synthetic videos, covering varying temporal dynamics for faithfulness and diverse knowledge categories for factuality. Furthermore, it supports four evaluation modes: Base (I), Visual Degradation (II), Evidence Corruption (III), and Temporal Intervention (IV) for order-sensitive items.

The non-Base modes apply controlled video perturbations while keeping questions fixed. Modes II--III are invariant-label settings and are evaluated by Resist Rate (RR), which measures whether correct Base decisions remain stable under visual degradation or corrupted evidence.
Mode~IV disrupts the temporal structure required for correctness and is evaluated using Temporal Sensitivity Score (TSS), which measures whether the model’s predictions changes after shuffling or reversal.
Our evaluation reveals that reliability under induced conditions is not uniform: models tend to be more fragile when exposed to misleading evidence than to purely perceptual degradation.
Moreover, temporal interventions reveal that many models remain largely insensitive to order disruption in factuality questions, indicating to a gap in temporal grounding beyond clean-scenario accuracy.

Our contributions are three-fold:
\begin{enumerate}
      \item We introduce \textsc{INFACT}, a diagnostic benchmark comprising 9{,}800 QA instances spanning real and synthetic videos, with fine-grained taxonomies covering both faithfulness and factuality hallucinations.
    \item We propose a four-mode evaluation protocol (Base, Visual Degradation, Evidence Corruption, Temporal Intervention) with paired reliability metrics (RR and TSS) for measuring invariant-label stability and temporal sensitivity.
    \item  We conduct a systematic evaluation of 14 representative Video-LLMs, revealing their stability under invariant-label perturbations and temporal inertia on order-sensitive items.
\end{enumerate}

\begin{table*}[t]
    \centering
    \small
    \caption{\textbf{Comparison of INFACT with recent hallucination benchmarks for video understanding.}
    INFACT covers video-grounded faithfulness and world-knowledge-grounded factuality with fine-grained taxonomies.
    It also supports controlled evaluation modes for visual degradation, evidence corruption, and temporal intervention within a unified protocol.}

    \label{tab:benchmark_comparison}
    \setlength{\tabcolsep}{4pt}
    \renewcommand{\arraystretch}{1.1}
    \resizebox{\linewidth}{!}{%
    \begin{tabular}{l c c c c c c c c c c}
        \toprule
        \multirow{2}{*}[-1.5ex]{\textbf{Benchmark}} &
        \multirow{2}{*}[-1.5ex]{\makecell[c]{\textbf{\# Ques.} / \\ \textbf{\# Videos}}} &
        \multirow{2}{*}[-1.5ex]{\textbf{Faithfulness}} &
        \multirow{2}{*}[-1.5ex]{\textbf{Factuality}} &
        \multirow{2}{*}[-1.5ex]{\textbf{Source}} &
        \multicolumn{2}{c}{\textbf{Visual Degradation}} &
        \multicolumn{2}{c}{\textbf{Evidence Corruption}} &
        \multicolumn{2}{c}{\textbf{Temporal Intervention}} \\
        \cmidrule(lr){6-7}\cmidrule(lr){8-9}\cmidrule(lr){10-11}
        & & & & &
        \makecell{\textbf{Motion}\\\textbf{Blur}} &
        \makecell{\textbf{Gaussian}\\\textbf{Noise}} &
        \makecell{\textbf{Caption}\\\textbf{Injection}} &
        \makecell{\textbf{Adversarial}\\\textbf{Noise}} &
        \makecell{\textbf{Shuffle}} &
        \makecell{\textbf{Reverse}} \\

        \midrule

        VIDHAL\cite{choong2024vidhal}
            & -- / 400
            & \cmark (5) & \xmark
            & Real
            & \xmark & \xmark
            & \xmark & \xmark
            & \xmark & \xmark \\

        EventHallusion\cite{zhang2024eventhallusion}
            & -- / 400
            & \cmark (3) & \xmark
            & Real
            & \xmark & \xmark
            & \xmark & \xmark
            & \xmark & \xmark \\

        VideoHallucer\cite{wang2024videohallucer}
            & 1{,}800 / 948
            & \cmark (3) & \cmark (3)
            & Real
            & \xmark & \xmark
            & \xmark & \xmark
            & \xmark & \xmark \\

        VIDHALLUC\cite{vidhalluc}
            & 9{,}295 / 5{,}002
            & \cmark (3) & \xmark
            & Real
            & \xmark & \xmark
            & \xmark & \xmark
            & \xmark & \xmark \\

        VideoHallu\cite{li2025videohallu}
            & 3{,}233 / 3{,}233
            & \cmark (2) & \cmark (4)
            & Synthetic
            & \xmark & \xmark
            & \xmark & \xmark
            & \xmark & \xmark \\

        \ours
            & 9{,}800 / 9{,}800
            & \cmark (12) & \cmark (12)
            & Real \& Synthetic
            & \cmark & \cmark
            & \cmark & \cmark
            & \cmark & \cmark \\
        \bottomrule
    \end{tabular}}
\end{table*}

\label{sec:data_construction}

\section{Related Works}

\subsection{Hallucination evaluation in Video-LLMs}

A growing line of work has benchmarked hallucinations in Video-LLMs from different perspectives.
Existing benchmarks~\cite{zhang2024eventhallusion, videomme2025_CVPR, vidhalluc, Bae_2025_CVP_MASH_VLM, ICLR2025_3cc68578_avhbench, wang2024videohallucer, Kong_Zeng_Chen_Li_Yan_Zhu_2025_mhbench} predominantly focus on faithfulness hallucinations, typically through controlled benchmark constructions that manipulate events, motion, video similarity, or cross-modal consistency to test grounding in the input video.
For instance, EventHallusion~\cite{zhang2024eventhallusion} focuses on event-level dynamics and relations, MHBench~\cite{Kong_Zeng_Chen_Li_Yan_Zhu_2025_mhbench} targets motion-related errors with adversarial triplets, and VidHalluc~\cite{vidhalluc} constructs visually distinct yet semantically similar video pairs to expose fragile grounding.
Related controlled settings further investigate shortcut behaviors driven by priors or spurious correlations, such as narrative priors in NOAH~\cite{lee2025noahbenchmarkingnarrativeprior}, action-scene correlations in UNSCENE~\cite{Bae_2025_CVP_MASH_VLM}, and cross-modal inconsistency settings in AVHBench~\cite{ICLR2025_3cc68578_avhbench}.
Some benchmarks broaden data sources or verification scope: VideoHallu~\cite{li2025videohallu} introduces synthetic videos, while VidHallucer~\cite{wang2024videohallucer} distinguishes cases by whether the target claim can be verified from the video.

However, as summarized in Table~\ref{tab:benchmark_comparison}, most existing benchmarks concentrate on hallucinations that can be judged against the input video (faithfulness), whereas factuality hallucinations requiring verifiable world knowledge are much less explored.

\subsection{Existing Video Understanding Benchmarks}

Existing video understanding benchmarks primarily evaluate Video-LLMs in clean settings with capability-oriented scoring. Broad-spectrum suites such as MVBench~\cite{mvbench2024}, Video-MME~\cite{videomme2025_CVPR}, and TOMATO~\cite{ICLR2025_tomato} target general task coverage, while ViMUL-Bench~\cite{shafique2025culturallydiversemultilingualmultimodalvideo} and CityGuesser-style QA~\cite{eccv_cityguessr} emphasize knowledge-heavy settings.

However, clean-input capability scores do not directly measure reliability: high accuracy can mask shortcut-based success, in which models rely on language priors, static cues, or dataset biases rather than video-dependent evidence~\cite{liu2025phd,yan2025shale,timecompass,Bae_2025_CVP_MASH_VLM}.

To reduce shortcut effects, several benchmarks adopt controlled designs such as conflicting videos or temporal multiple-choice setups~\cite{timecompass,cores2025losttimenewtemporal_tvbench}. Although useful for diagnosing temporal dependence, these protocols are not formulated as hallucination evaluations: they do not separate evidence bases (video vs.\ verifiable world knowledge) and do not probe reliability under explicit evidence-corruption conditions.

\section{INFACT}

In this section, we introduce INFACT, a fine-grained benchmark designed to evaluate Video-LLMs faithfulness and factuality in both clean and noisy scenarios. We first detail the taxonomy of hallucinations in \S~\ref{Taxonomy_section} and the data construction process in \S~\ref{sec:data_construction_section}, describing the aggregation of 9{,}800 questions from public video-QA datasets, instructional resources, and synthetic collections. To investigate the root causes of model failure, \S~\ref{ref:Hallucination_induction_designs} introduces three hallucination induction modes devised to probe reliability under visual degradation, evidence corruption, and temporal intervention. Finally, we introduce two kinds of evaluation metrics to quantify reliability in \S~\ref{sec:metrics}.

\begin{figure*}[t]
  \centering
  \includegraphics[width=\linewidth,
    trim=8mm 5mm 1mm 10mm, 
    clip]{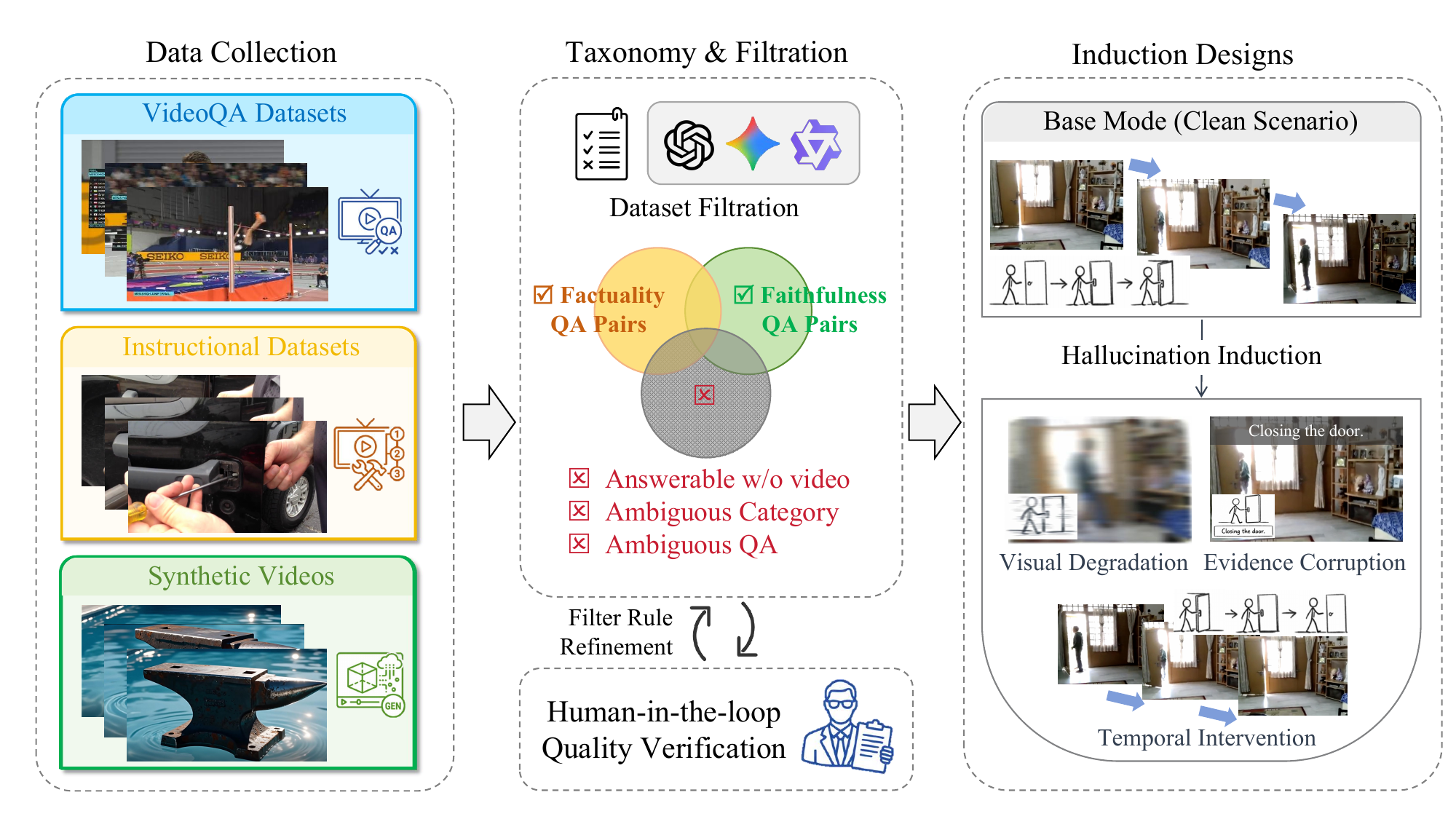}
\caption{\textbf{Overview of the INFACT construction process.}
\textbf{Left:} Candidate videos and QA pairs are collected from multiple sources, including video QA datasets, instructional datasets, and synthetic videos.
\textbf{Middle:} Samples are organized into fine-grained faithfulness and factuality dimensions, and filtered to remove ambiguous or non-video-grounded items, followed by human-in-the-loop quality verification.
\textbf{Right:} The resulting benchmark supports four evaluation modes: Base, Visual Degradation, Evidence Corruption, and Temporal Intervention.}
  \label{fig:framework}
\end{figure*}

\subsection{Taxonomy}
We categorize hallucinations by the evidentiary basis required for verification, where faithfulness requires alignment with visual content and factuality necessitates consistency with world knowledge.
\label{Taxonomy_section}

\paragraph{Faithfulness }
hallucinations occur when model outputs contradict explicit visual evidence in the video.
Following the common object/attribute/relation hierarchy ~\cite{liu2024surveyhallucinationlargevisionlanguage} used in static vision-language evaluation, we extend this framework to videos by incorporating temporal dynamics. This yields three hierarchical levels organized by increasing spatiotemporal complexity (Table~\ref{tab:faithful_taxonomy}).
\textbf{Level 1 (Static Entities \& Attributes)} corresponds to \emph{Object} and \emph{Attribute} tiers. This level requires local perception to resolve \emph{Entity Recognition}, \emph{Unique Entity Counting}, \emph{Temporal Attributes Recognition}, \emph{Static Attributes Recognition}, and \emph{Scene Text Recognition}.
\textbf{Level 2 (Dynamic Actions \& Motions)} requires dynamic perception to aggregate visual features over time to resolve \emph{Action Recognition}, \emph{Repetitive Action Counting} and \emph{Motion Attributes Recognition}.
\textbf{Level 3 (Spatio-Temporal Relations)} corresponds to relation-level reasoning based on global spatio-temporal structure, including \emph{Spatial Relation Recognition}, \emph{Temporal Relation Recognition}, \emph{State Transition Detection}, and \emph{Temporal Localization}.

\begin{figure}[thb]
    \centering
    \includegraphics[width=\linewidth,
    trim=5mm 42mm 5mm 40mm, 
    clip]{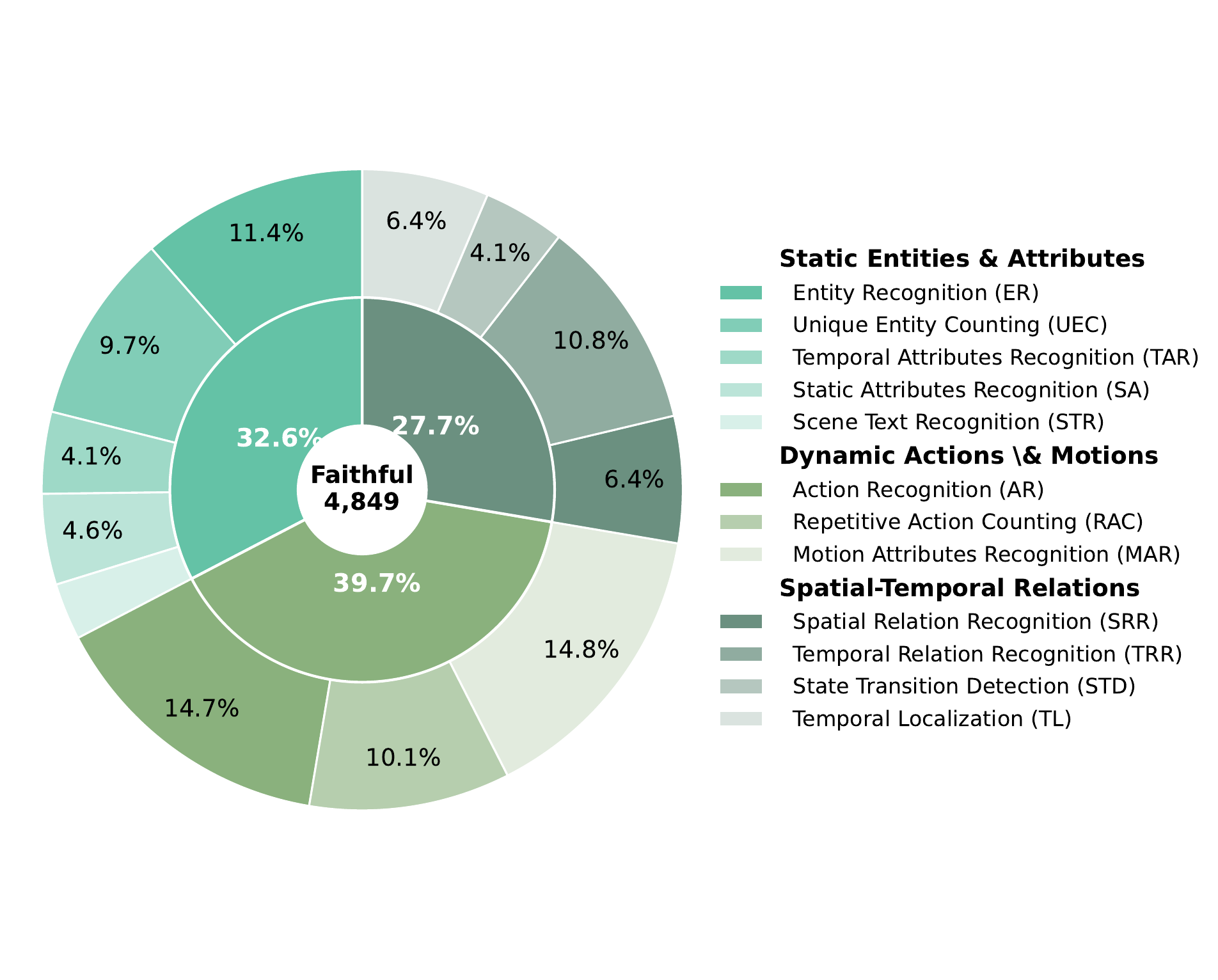}
    \vspace{0mm} 
    \includegraphics[width=\linewidth,
    trim=5mm 25mm 5mm 25mm,
    clip]{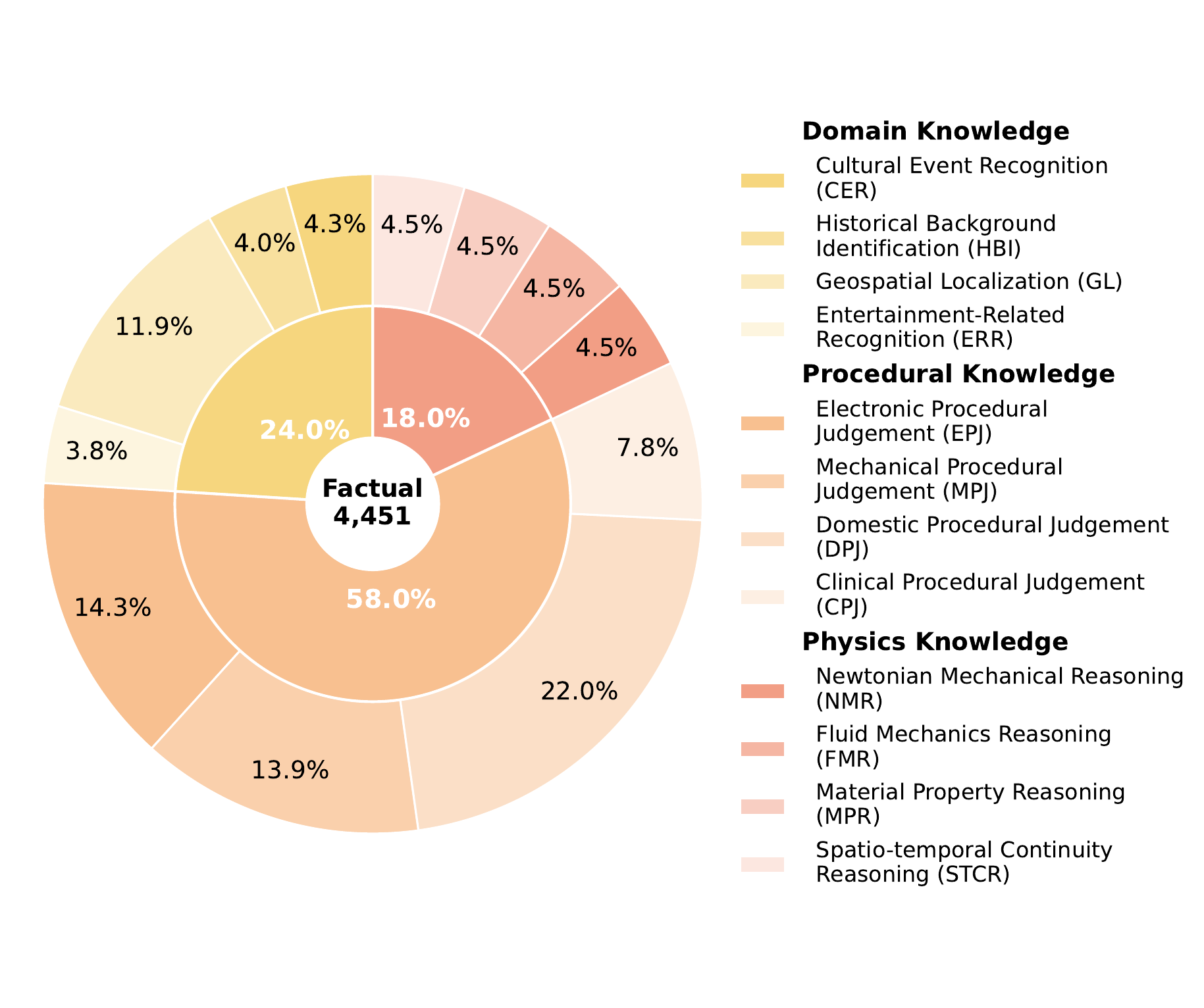}

    \caption{\textbf{Dataset composition of INFACT.}
    Distribution over the fine-grained taxonomy for faithfulness (top) and factuality (bottom).}
        
    \label{fig:statistic}
\end{figure}

\paragraph{Factuality}
hallucinations occur when model outputs contradict world knowledge, requiring information beyond what is present in the video content alone. We structure factuality into three categories based on the knowledge required (Table~\ref{tab:factual_taxonomy}).
\textbf{Domain Knowledge (Know-WHAT)} evaluates consistency with verifiable world knowledge across diverse domains, including \emph{Cultural Event Recognition}, \emph{Historical Background Identification}, \emph{Geospatial Localization}, and \emph{Entertainment-Related Recognition}.
\textbf{Procedural Knowledge (Know-HOW)} assesses instructional validity by verifying whether the described sequence of steps respects prerequisite relations and causal dependencies. This category covers \emph{Electronic}, \emph{Mechanical}, \emph{Domestic}, and \emph{Clinical} domains.
\textbf{Physical Knowledge (Know-WHY)} probes the adherence of a model to fundamental physical laws, including \emph{Mechanical}, \emph{Fluid Mechanics}, \emph{Material Property}, and \emph{Spatial-Temporal continuity}. 
To ensure rigor, We restrict factuality to objectively verifiable knowledge and explicitly exclude subjective or ambiguous claims.
Figure~\ref{fig:statistic} summarizes the distribution of samples across the fine-grained taxonomy for both faithfulness and factuality.

\subsection{Data Construction}
\label{sec:data_construction_section}
\paragraph{Faithfulness Data.}

To construct a rigorous benchmark for hallucination evaluation, we curate high-quality samples from established video understanding benchmarks, including MVBench~\cite{mvbench2024}, Video-MME~\cite{videomme2025_CVPR}, and TOMATO~\cite{ICLR2025_tomato}, which provide diverse questions suitable for video evidence verification.
To guarantee data quality, we employ a two-stage alignment process.
First, we manually map the categories of source benchmarks to our proposed taxonomy. Subsequently, we implement a "taxonomy-aligned filtration" pipeline that utilizes LLM-assisted consensus labeling. This ensures strict adherence to the spatio-temporal complexity levels defined in Appendix  Table~\ref{tab:faithful_taxonomy}, thereby refining the quality of constructed benchmark.

\paragraph{Factuality Data.}
Factuality samples are constructed to cover three verifiable knowledge requirements: Domain, Procedural, and Physical Knowledge.
For \textbf{Domain Knowledge}, candidates are drawn from knowledge-intensive video QA resources such as CityGuessr68k~\cite{eccv_cityguessr} and ViMULBench~\cite{shafique2025culturallydiversemultilingualmultimodalvideo} to ensure comprehensive coverage across diverse domains such as culture, history, geography, and entertainment.
For \textbf{Procedural Knowledge}, instructional videos are sourced from COIN~\cite{coindataset} and MedVidQA~\cite{gupta2023dataset}, where step annotations provide temporal boundaries for each procedure segment.
To create counterfactuals, we randomly shuffle these segments to disrupt the temporal order required by prerequisite relations and causal dependencies.
The resulting videos support \emph{binary} procedural judgement (correct vs.\ incorrect).
For \textbf{Physical Knowledge}, physically implausible videos are synthesized using text-to-video generation models (e.g., Sora~\cite{sora}, Wan2.5~\cite{wan25}, and Gemini Veo 3~\cite{geminiveo3}). These samples violate physical laws, such as gravity-defying motion, serving as a test for physical reasoning capabilities.

\paragraph{Dataset Filtration and Quality Review.}
An LLM-ensemble filter is first applied to remove samples with linguistic ambiguity and those solvable without video evidence (e.g., answerable from text-only priors).
Since source annotations and edge cases in taxonomy mapping may still introduce noise, a \textbf{human-in-the-loop} verification stage is further conducted.
Rather than discarding individual problematic samples only, annotators also identify recurring error patterns and trace them back to their causes (e.g., a question template or a mapping rule).
This enables iterative updates to the upstream prompts and filtering logic and batch correction of affected samples.
The process is repeated until the benchmark satisfies the quality criteria.

\subsection{Hallucination Induction Designs}
\label{ref:Hallucination_induction_designs}
INFACT establishes four distinct evaluation modes to probe evidence-consistent behavior under controlled conditions.
Mode~I (Base) establishes the baseline performance on clean data.
Modes II \& III (Robustness) apply label-preserving transformations to test whether models maintain correct answers under visual degradation or conflicting evidence.
Mode~IV (Sensitivity) applies label-changing interventions to test sensitivity to disrupted temporal structure.

\paragraph{Mode I: Base (clean scenario).}
The Base mode evaluates the model on the original, unaltered video and question pairs.
It serves as the reference standard for establishing upper-bound performance.
Crucially, to isolate the impact of interventions from intrinsic item difficulty, we employ a \emph{paired comparison protocol}: behavior under all induced modes is assessed relative to the model's Base outcome on the same specific item.

\paragraph{Mode II: Visual Degradation.}
This mode probes hallucinations induced by perceptual uncertainty.
We introduce three types of visual perturbations: (1) \textbf{Gaussian noise} and (2) \textbf{Motion blur}, both applied uniformly at the frame level, and (3) \textbf{Video Compression}, which simulates platform-specific re-encoding artifacts by reducing the bitrate to a lossy threshold.
While these operators reduce low-level perceptual clarity, they preserve the high-level semantic content required to answer the question.
Consequently, a reliable model should maintain the correct answer despite the visual noise, rather than introducing unsupported details when the evidence becomes harder to perceive.

\paragraph{Mode III: Evidence Corruption.}
Since Video-LLMs often integrate auxiliary textual signals (e.g, subtitles) with visual data, reliability hinges on the ability to prioritize authentic visual evidence over misleading external cues.
We employ three operators to simulate untrusted conditioning while keeping the ground-truth label invariant.
\textbf{Caption injection} introduces a cross-modal conflict by overlaying subtitles on randomly sampled video segments. The injected subtitles mix (i) content-irrelevant sentences and (ii) LLM-generated misleading statements conditioned on the question and the ground-truth answer option, constructed to form a plausible but incorrect textual cue that conflicts with the video evidence required for correctness (e.g., a video showing \emph{opening a door} is paired with the subtitle \emph{closing the door}).
\textbf{Subtitle corruption} injects noisy ASR-like subtitles with subtitle--video desynchronization, simulating the unreliable OCR/ASR outputs that models may encounter when processing captioned or subtitled content in the wild.
\textbf{Adversarial noise} targets the visual encoder by applying a transfer-based black-box perturbation generated with MI-FGSM~\cite{Dong_2018_CVPR_MIFGSM}. 
To ensure the attack generalizes across different architectures, perturbations are produced using a proxy ensemble of visual encoders (InternVL3-8B~\cite{zhu2025internvl3}, Qwen3VL-8B~\cite{Qwen3-VL}, and Video-MAE~\cite{tong2022videomae}) to reduce reliance on a single proxy model.
Since the ground-truth label is intended to remain unchanged, the desired behavior is \emph{stability} against misleading conditioning.

\paragraph{Mode IV: Temporal Intervention.}
This mode probes temporal evidence consistency.
Specifically, whether the model's correctness genuinely stems from understanding event order and state transitions.
Unlike the previous modes, here the temporal structure is intentionally destroyed, rendering the original ground truth invalid. 
To evaluate this, we curate an \emph{order-sensitive} subset from Action Dynamics, Spatio-Temporal Structure, and Procedural Knowledge, where the chronological sequence is strictly required for correctness.
We apply temporal intervention to these videos via frame-level shuffling or reversal.
If a model retains the Base prediction after intervention, it indicates temporal insensitivity and reliance on order-invariant cues.

\subsection{Evaluation Metrics}
\label{sec:metrics}

Accuracy is reported under the Base (clean-scenario) setting to measure performance on faithfulness and factuality. To quantify reliability under induced conditions, we additionally report metrics tailored to invariant-label settings and temporal interventions. Modes II–III (Visual Degradation and Evidence Corruption) are evaluated by Resist Rate (RR), while Mode IV (Temporal Intervention) is evaluated by Temporal Sensitivity Score (TSS).

\paragraph{Resist Rate (RR).}
For visual degradations ($\mathcal{T}_{\text{deg}}$) and evidence corruptions ($\mathcal{T}_{\text{cor}}$), the ground-truth label is intended to remain unchanged.
We measure reliability by whether a model preserves its correct Base prediction under each operator.
For an operator $p \in \mathcal{T}_{\text{pert}}=\mathcal{T}_{\text{deg}} \cup \mathcal{T}_{\text{cor}}$, we define:
\begin{equation*}
\resizebox{1\hsize}{!}{$
\mathrm{RR}_{p} =
\frac{\sum_{i} \mathbb{I}\big( f(V_i, q_i) = y_i^{\mathrm{gt}} \big) \cdot \mathbb{I}\big( f(p(V_i), q_i) = y_i^{\mathrm{gt}} \big)}
{\sum_{i} \mathbb{I}\big( f(V_i, q_i) = y_i^{\mathrm{gt}} \big)},
$}
\end{equation*}
where $(V_i, q_i)$ denotes the $i$-th video-question pair, $y_i^{\mathrm{gt}}$ is the corresponding ground-truth answer, $f(\cdot,\cdot)$ denotes a Video-LLM, and $\mathbb{I}(\cdot)$ is the indicator function.
We report operator-wise scores (e.g., $RR_{\mathrm{gau}}, RR_{\mathrm{mb}}, RR_{\mathrm{adv}}, RR_{\mathrm{cap}}$), and compute
$RR_{\text{deg}}$ and $RR_{\text{cor}}$ as the mean over degradation and corruption operators, respectively.

\begin{table*}[tbh]
    \centering
    \small
    \caption{\textbf{Faithfulness results on INFACT.}
    Text-only: question-only accuracy. Base: Mode~I. RR: Modes~II--III. TSS: Mode~IV (order-sensitive subset).
    Models are grouped by availability and sorted by Avg Score.}
    \label{tab:faithful}
    \setlength{\tabcolsep}{2pt}
    \renewcommand{\arraystretch}{1.05}
    \resizebox{\linewidth}{!}{%
\begin{tabular}{lcccccccccccccc}
    \toprule
    & & & \multicolumn{4}{c}{Evidence Corruption} & \multicolumn{4}{c}{Visual Degradation} & \multicolumn{3}{c}{Temporal Intervention} & \multicolumn{1}{c}{}\\
    \cmidrule(lr){4-7} \cmidrule(lr){8-11} \cmidrule(lr){12-14}
    Model & Text-only & Base & $RR_{\mathrm{adv}}$ & $RR_{\mathrm{cap}}$ & $RR_{\mathrm{sub}}$ & $RR_{\mathrm{cor}}$ & $RR_{\mathrm{cmp}}$ & $RR_{\mathrm{gau}}$ & $RR_{\mathrm{mb}}$ & $RR_{\mathrm{deg}}$ & $TSS_{\mathrm{shu}}$ & $TSS_{\mathrm{rev}}$ & $\bar{\mathrm{TSS}}$ & Avg Score \\
    \midrule
    PLLaVA-13B \cite{xu2024pllava} & 0.285 & 0.458 & 0.782 & 0.678 & 0.922 & 0.794 & 0.777 & 0.813 & 0.699 & 0.763 & 0.010 & 0.014 & 0.012 & 0.523 \\
    VideoLLaMA2-7B\cite{damonlpsg2024videollama2} & 0.261 & 0.434 & 0.751 & 0.564 & 0.910 & 0.742 & 0.818 & 0.725 & 0.749 & 0.764 & 0.113 & 0.091 & 0.102 & 0.536 \\
    ShareGPT4Video-8B \cite{chen2024sharegpt4videoimprovingvideounderstanding} & 0.292 & 0.462 & 0.719 & 0.688 & 0.894 & 0.767 & 0.840 & 0.762 & 0.754 & 0.785 & 0.101 & 0.094 & 0.098 & 0.550 \\
    PLLaVA-34B \cite{xu2024pllava} & 0.287 & 0.505 & 0.757 & 0.618 & 0.920 & 0.765 & 0.772 & \best{0.874} & 0.848 & \second{0.831} & 0.070 & 0.073 & 0.072 & 0.556 \\
    Tarsier-34B \cite{wang2024tarsierrecipestrainingevaluating} & \second{0.302} & 0.568 & 0.785 & 0.644 & 0.960 & 0.796 & 0.759 & 0.846 & 0.612 & 0.739 & 0.158 & 0.208 & 0.183 & 0.573 \\
    Qwen2.5VL-7B\cite{Qwen2.5-VL} & 0.278 & 0.531 & 0.630 & 0.659 & 0.973 & 0.754 & 0.804 & 0.727 & 0.640 & 0.724 & 0.230 & 0.258 & 0.244 & 0.574 \\
    NVILA-8B \cite{liu2024NVILA} & 0.293 & 0.533 & 0.804 & 0.669 & 0.963 & 0.812 & 0.797 & 0.853 & 0.809 & 0.820 & 0.122 & 0.150 & 0.136 & 0.589 \\
    Qwen3VL-8B\cite{Qwen3-VL} & 0.295 & 0.557 & 0.676 & 0.715 & 0.941 & 0.777 & 0.829 & 0.785 & 0.744 & 0.786 & 0.224 & 0.189 & 0.207 & 0.590 \\
    Qwen2.5VL-32B\cite{Qwen2.5-VL} & 0.286 & 0.538 & \second{0.817} & \second{0.724} & 0.949 & 0.830 & 0.790 & 0.824 & 0.773 & 0.796 & 0.198 & 0.204 & 0.201 & 0.609 \\
    InternVL3-8B \cite{zhu2025internvl3} & 0.301 & 0.576 & 0.777 & 0.709 & \best{0.979} & 0.822 & 0.828 & 0.847 & 0.817 & \second{0.831} & 0.169 & 0.212 & 0.191 & 0.615 \\
    InternVL3.5-8B \cite{wang2025internvl35} & 0.294 & \second{0.606} & 0.796 & 0.721 & \second{0.978} & \second{0.832} & 0.841 & 0.792 & \second{0.850} & 0.828 & 0.171 & 0.241 & 0.206 & 0.622 \\
    Qwen3VL-32B \cite{Qwen3-VL} & 0.287 & 0.602 & 0.809 & 0.718 & 0.975 & 0.834 & 0.802 & 0.836 & 0.815 & 0.818 & \second{0.248} & \second{0.289} & \second{0.269} & \second{0.640} \\
    GPT-5.1 \cite{openai_gpt51_2025} & \best{0.305} & \second{0.687} & \second{0.856} & \second{0.823} & 0.958 & \second{0.879} & \second{0.940} & 0.801 & \second{0.854} & \second{0.865} & \second{0.355} & 0.281 & \second{0.318} & \second{0.687} \\
    Gemini3-flash\cite{gemini3flash} & 0.295 & \best{0.784} & \best{0.886} & \best{0.914} & 0.962 & \best{0.921} & \best{0.942} & \second{0.859} & \best{0.892} & \best{0.898} & \best{0.581} & \best{0.492} & \best{0.537} & \best{0.785} \\
    \bottomrule
    \end{tabular}}
\end{table*}

\begin{table*}[tbh]
\centering
\small
\caption{\textbf{Factuality results on INFACT.} Same evaluation protocol and metrics as Table~\ref{tab:faithful}.}
\label{tab:factual}
\setlength{\tabcolsep}{2pt}
\renewcommand{\arraystretch}{1.05}
\resizebox{\linewidth}{!}{%
\begin{tabular}{lcccccccccccccc}
\toprule
& & & \multicolumn{4}{c}{Evidence Corruption} & \multicolumn{4}{c}{Visual Degradation} & \multicolumn{3}{c}{Temporal Intervention} & \multicolumn{1}{c}{} \\
\cmidrule(lr){4-7} \cmidrule(lr){8-11} \cmidrule(lr){12-14}
Model & Text-only & Base & $RR_{\mathrm{adv}}$ & $RR_{\mathrm{cap}}$ & $RR_{\mathrm{sub}}$ & $RR_{\mathrm{cor}}$ & $RR_{\mathrm{cmp}}$ & $RR_{\mathrm{gau}}$ & $RR_{\mathrm{mb}}$ & $RR_{\mathrm{deg}}$ & $TSS_{\mathrm{shu}}$ & $TSS_{\mathrm{rev}}$ & $\bar{\mathrm{TSS}}$ & Avg Score \\
\midrule
VideoLLaMA2-7B\cite{damonlpsg2024videollama2} & 0.272 & 0.407 & 0.676 & 0.590 & 0.917 & 0.728 & 0.818 & 0.723 & 0.772 & 0.771 & 0.000 & 0.000 & 0.000 & 0.500 \\
ShareGPT4Video-8B \cite{chen2024sharegpt4videoimprovingvideounderstanding} & 0.294 & 0.489 & 0.715 & 0.519 & 0.883 & 0.706 & 0.840 & 0.816 & 0.798 & 0.818 & 0.000 & 0.000 & 0.000 & 0.508 \\
PLLaVA-13B \cite{xu2024pllava} & 0.279 & 0.410 & 0.788 & 0.585 & 0.919 & 0.764 & 0.777 & \best{0.947} & 0.705 & 0.810 & 0.000 & 0.000 & 0.000 & 0.525 \\
NVILA-8B \cite{liu2024NVILA} & 0.291 & 0.424 & 0.818 & 0.541 & 0.961 & 0.773 & 0.797 & 0.888 & 0.759 & 0.815 & 0.000 & 0.000 & 0.000 & 0.529 \\
PLLaVA-34B \cite{xu2024pllava} & 0.265 & 0.431 & 0.782 & 0.603 & 0.920 & 0.768 & 0.772 & 0.897 & 0.812 & 0.827 & 0.000 & 0.000 & 0.000 & 0.532 \\
InternVL3-8B \cite{zhu2025internvl3} & 0.295 & 0.465 & 0.738 & 0.612 & 0.951 & 0.767 & 0.828 & 0.914 & 0.813 & \second{0.852} & 0.000 & 0.000 & 0.000 & 0.540 \\
InternVL3.5-8B\cite{wang2025internvl35} & 0.298 & 0.498 & 0.775 & 0.619 & 0.955 & 0.783 & 0.841 & 0.921 & 0.851 & \best{0.871} & 0.000 & 0.000 & 0.000 & 0.551 \\
Tarsier-34B \cite{wang2024tarsierrecipestrainingevaluating} & 0.289 & 0.441 & \second{0.909} & \second{0.625} & 0.960 & \second{0.831} & 0.759 & \second{0.930} & 0.820 & 0.836 & 0.000 & 0.000 & 0.000 & 0.556 \\
Qwen2.5VL-32B\cite{Qwen2.5-VL} & 0.296 & 0.512 & 0.804 & 0.624 & 0.930 & 0.786 & 0.790 & 0.889 & 0.764 & 0.814 & 0.105 & 0.101 & 0.103 & 0.568 \\
Qwen2.5VL-7B\cite{Qwen2.5-VL} & 0.281 & 0.503 & 0.670 & 0.575 & \second{0.968} & 0.738 & 0.804 & 0.784 & 0.774 & 0.787 & 0.212 & 0.262 & 0.237 & 0.587 \\
Qwen3VL-8B \cite{Qwen3-VL} & 0.296 & 0.541 & 0.707 & 0.606 & 0.929 & 0.747 & 0.829 & 0.793 & 0.768 & 0.797 & \second{0.339} & \second{0.334} & \second{0.337} & 0.627 \\
Qwen3VL-32B \cite{Qwen3-VL} & \second{0.305} & 0.521 & 0.809 & 0.641 & \second{0.963} & 0.804 & 0.802 & 0.904 & 0.821 & 0.842 & 0.231 & 0.254 & 0.243 & 0.630 \\
GPT-5.1 \cite{openai_gpt51_2025} & \best{0.308} & \second{0.678} & 0.901 & \second{0.814} & \second{0.963} & \second{0.893} & \second{0.940} & 0.911 & \second{0.876} & \second{0.909} & \second{0.391} & \second{0.414} & \second{0.402} & \second{0.735} \\
Gemini3-flash\cite{gemini3flash} & \second{0.305} & \best{0.752} & \best{0.917} & \best{0.839} & \best{0.969} & \best{0.908} & \best{0.942} & 0.902 & \best{0.881} & 0.908 & \best{0.741} & \best{0.696} & \best{0.719} & \best{0.845} \\
\bottomrule
\end{tabular}}
\end{table*}

\begin{figure}[tbh]
  \centering
  \includegraphics[width=\linewidth, clip, trim=2mm 2mm 1mm 1mm]{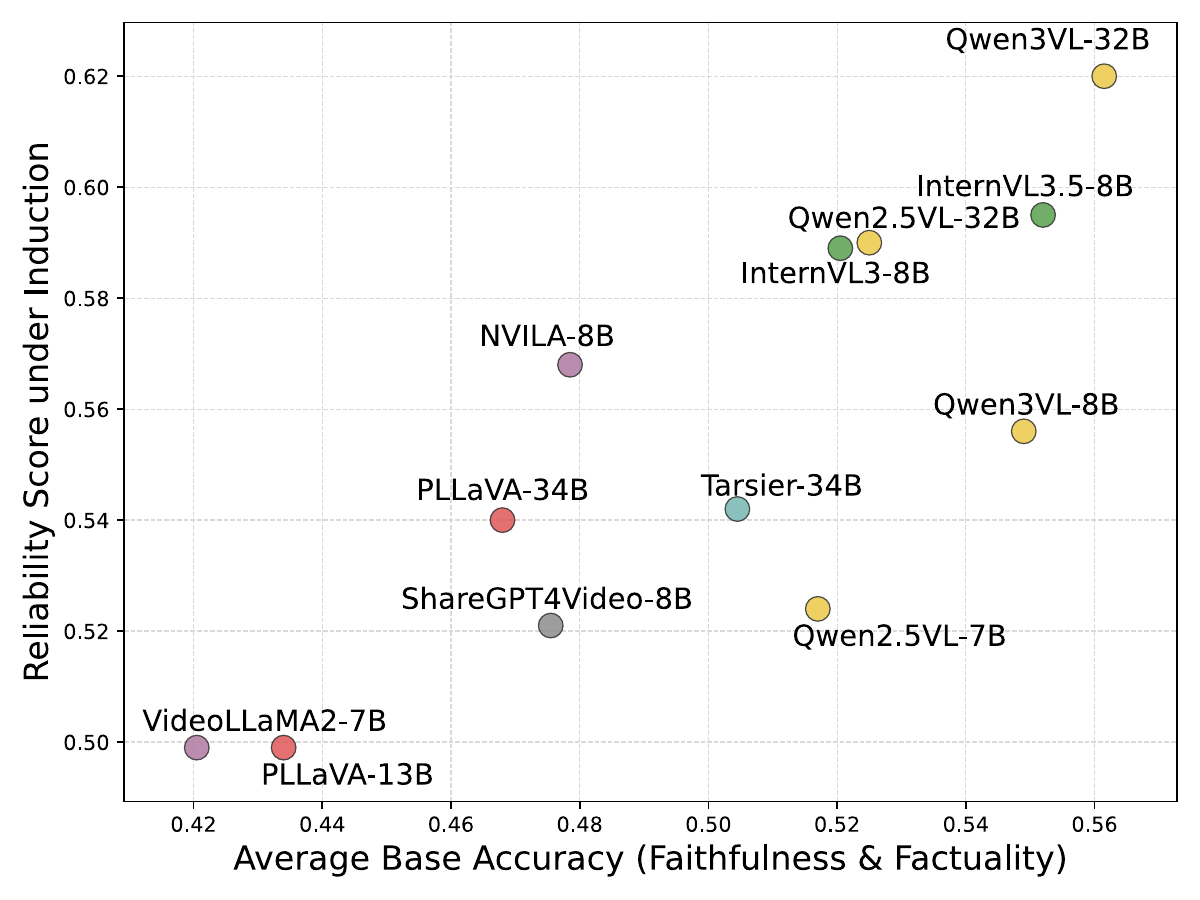}
    \caption{\textbf{Base accuracy vs.\ average reliability score under inductions.}
    Base accuracy is measured in Mode~I and averaged over faithfulness and factuality.
    The average reliability score under induction aggregates RR over Modes~II--III and TSS over Mode~IV.}
  \label{fig:clean_reliability}
\end{figure}

\begin{figure}[!t]
  \centering
  \includegraphics[width=\linewidth]{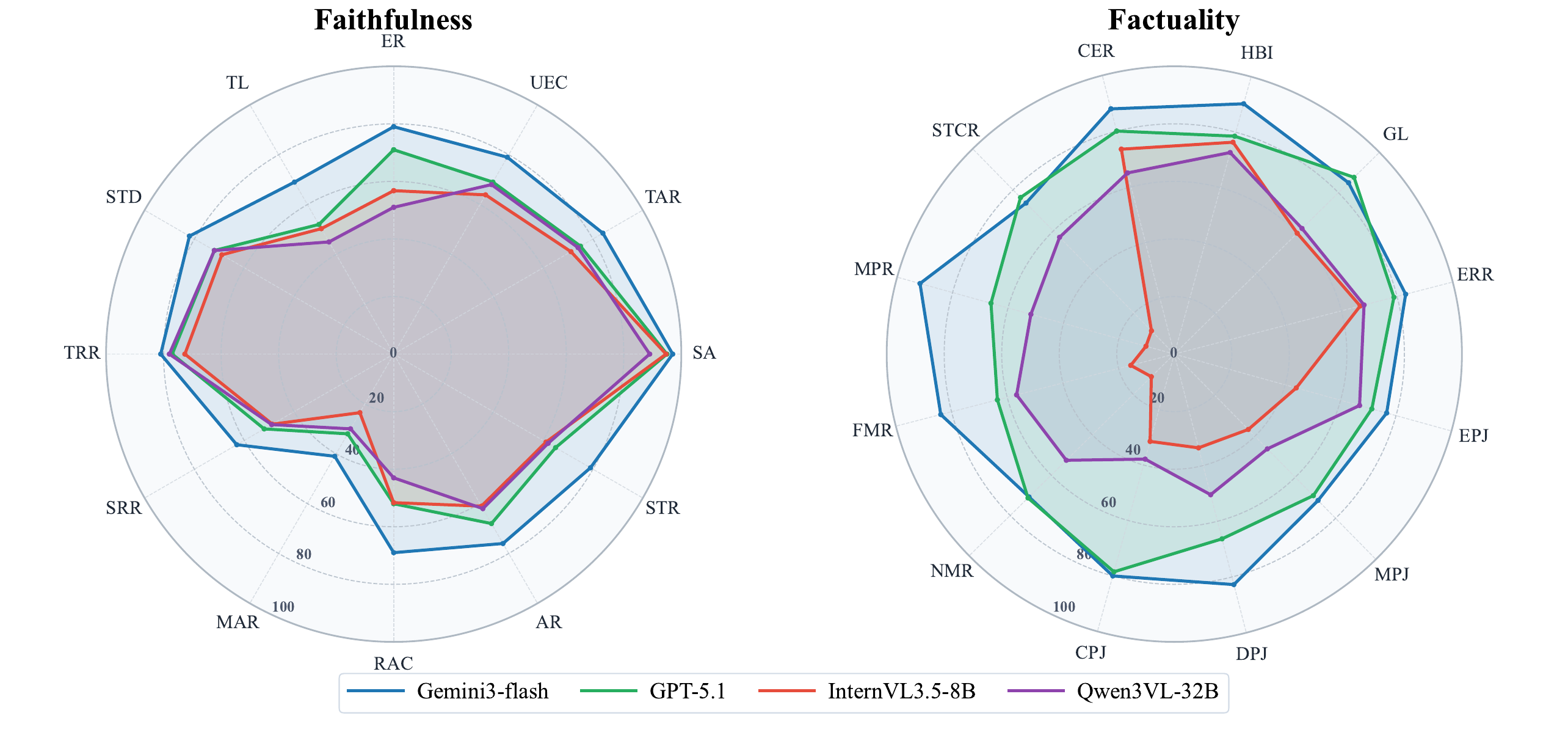}
\caption{\textbf{Comparison of four representative models on fine-grained evaluation dimensions.}
The left radar plot shows performance on faithfulness dimensions, while the right radar plot shows performance on factuality dimensions.
Each axis corresponds to a fine-grained category in the INFACT taxonomy, and each curve represents one representative model.
Higher values indicate better performance on the corresponding dimension.}
    \label{fig:radar}
\end{figure}

\begin{figure*}
    \centering
    \includegraphics[width=1\linewidth]{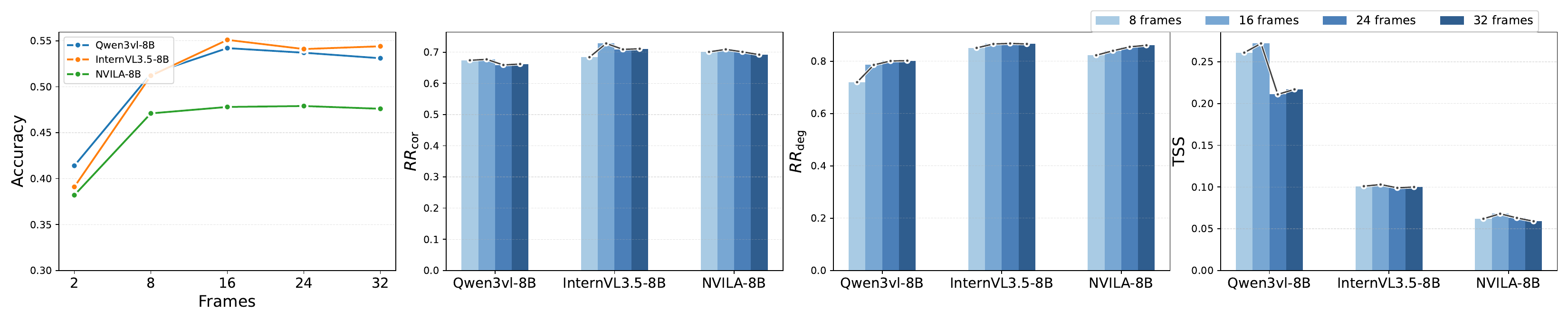}
    \caption{\textbf{Effect of the number of sampled frames.}
    \textbf{Left:} Base accuracy under $\{2,8,16,24,32\}$ uniformly sampled frames.
    \textbf{Right:} induced metrics ($RR_{\mathrm{deg}}$, $RR_{\mathrm{cor}}$, TSS) under $\{8,16,24,32\}$ frames.}
    \label{fig:frame}
\end{figure*}
\vspace{-12pt}

\paragraph{Temporal Sensitivity Score (TSS).}
Temporal interventions (shuffling or reversal) test whether model decisions are genuinely grounded in temporal structure.
TSS is computed on the \textbf{order-sensitive subset} $\mathcal{S}_{\text{order}}$, where temporal order is essential for correctness. Unlike the robustness tests used for RR, these interventions effectively invalidate the original ground-truth label. Consequently, a temporally grounded model should diverge from its initial decision after intervention. We define TSS as the rate at which model ceases to predict the original ground-truth label $y_{gt}$ after intervention.
For an intervention operator $p \in \mathcal{T}_{\text{iv}}$ (e.g., shuffling or reversal), we define:
\begin{equation*}
\resizebox{1\hsize}{!}{$
\mathrm{TSS}_{p} =
\frac{\sum_{i \in \mathcal{S}_{\text{order}}} \mathbb{I}(f(V_i, q_i)=y_{\mathrm{gt}}) \cdot \mathbb{I}(f(p(V_i), q_i)\neq y_{\mathrm{gt}})}
{\sum_{V_i \in \mathcal{S}_{\text{order}}} \mathbb{I}(f(V_i, q_i)=y_{\mathrm{gt}})}.
$}
\end{equation*}
We report $\mathrm{TSS}_{\mathrm{shu}}$ and $\mathrm{TSS}_{\mathrm{rev}}$, and $\bar{\mathrm{TSS}}$ is their mean.
A low TSS indicates \emph{temporal inertia}: the model tends to retain the Base decision even when the supporting temporal evidence is destroyed, suggesting a reliance on static priors rather than temporal logic.

\section{Experiments}

\subsection{Setups}

Fourteen models are evaluated on INFACT, including two proprietary systems (GPT-5.1 ~\cite{openai_gpt51_2025}, Gemini3-flash~\cite{gemini3flash}) and twelve open-source baselines: VideoLLaMA2-7B~\cite{damonlpsg2024videollama2}, PLLaVA-13B/34B~\cite{xu2024pllava}, ShareGPT4Video-8B~\cite{chen2024sharegpt4videoimprovingvideounderstanding}, Qwen2.5-VL-7B/32B~\cite{Qwen2.5-VL}, Qwen3-VL-8B/32B~\cite{Qwen3-VL}, Tarsier-34B~\cite{wang2024tarsierrecipestrainingevaluating}, NVILA-8B~\cite{liu2024NVILA}, InternVL3-8B~\cite{zhu2025internvl3}, and InternVL3.5-8B~\cite{wang2025internvl35}.
All models are evaluated zero-shot with default prompts using 16 uniformly sampled frames per video~\cite{ICLR2025_tomato,rawal2025argushallucinationomissionevaluation}.

\subsection{Results and Analysis}
\paragraph{Base accuracy vs. reliability under induced modes.}
Figure~\ref{fig:clean_reliability} reveals a strong association between Base accuracy (Mode I) and induced-mode reliability (Modes II--IV), averaged over faithfulness and factuality (Pearson $r{=}0.978$; Spearman $\rho{=}0.969$).
However, rankings are not fully preserved: models with similar Base accuracy can diverge under controlled perturbations.
Tables~\ref{tab:faithful}--\ref{tab:factual} suggest that these gaps are driven by differences in stability (RR) and temporal sensitivity (TSS), so clean accuracy alone is insufficient to characterize evidence-consistent behavior.

\paragraph{Stability under Visual Degradation \& Evidence Corruption Inductions (RR).}
Tables~\ref{tab:faithful}--\ref{tab:factual} show that evidence corruption degrades RR more than visual degradation, and caption injection is typically more damaging than transfer-based adversarial noise.
For instance, VideoLLaMA2-7B drops from $RR_{\mathrm{adv}}{=}0.751$ to $RR_{\mathrm{cap}}{=}0.564$ on faithfulness, and Tarsier-34B drops from $RR_{\mathrm{adv}}{=}0.909$ to $RR_{\mathrm{cap}}{=}0.625$ on factuality.
This pattern suggests that stability is more fragile when misleading auxiliary cues are introduced than when the visual signal is merely degraded.

\paragraph{Sensitivity under Temporal Intervention (TSS).}
Tables~\ref{tab:faithful}--\ref{tab:factual} report TSS on the order-sensitive subset, where frame shuffling or reversal disrupts the temporal structure required for correctness.
Several open-source baselines exhibit \emph{temporal inertia}, most visibly on factuality where multiple models yield $\bar{\mathrm{TSS}}{=}0$, with predictions unchanged after shuffling or reversal on Base-correct order-sensitive items.
In contrast, Qwen3VL-8B and Gemini3-flash attain $0.337$ and $0.719$, indicating that temporal intervention separates models by their reliance on temporal evidence.
Among open-source baselines, Qwen2.5/3-VL achieves comparatively higher TSS (Tables~\ref{tab:faithful}--\ref{tab:factual}), highlighting the potential role of more explicit time-aligned spatiotemporal positional encoding (e.g., mRoPE-style designs).

\paragraph{Performance across Fine-grained Dimensions.}
For finer diagnosis, we further analyze four representative models across fine-grained faithfulness and factuality dimensions (Figure~\ref{fig:radar}).
Across these models, factuality dimensions tend to lag behind faithfulness dimensions, especially in procedural judgement and physical reasoning.
On the faithfulness side, the remaining weaknesses concentrate on motion- and structure-centric dimensions (e.g., temporal localization and motion attributes), suggesting that fine-grained temporal aggregation remains challenging under our evaluation setting.

\paragraph{Effect of Frame Sampling Rate.}
We conduct a frame-ablation study on 500 videos randomly sampled from INFACT (Figure~\ref{fig:frame}). Base accuracy is evaluated with $\{2,8,16,24,32\}$ uniformly sampled frames and quickly saturates, where 16 frames already capture most of the gain (Figure~\ref{fig:frame}, left). Under induced modes with $\{8,16,24,32\}$ frames, $RR_{\mathrm{deg}}$ and $RR_{\mathrm{cor}}$ vary modestly and TSS shows no consistent upward trend (Figure~\ref{fig:frame}, right), suggesting that increasing the number of sampled frames alone does not consistently improve reliability under induced modes within this range, particularly for temporal sensitivity.

\section{Conclusion}
We present \textsc{INFACT}, a fine-grained benchmark for evaluating faithfulness and factuality hallucinations in Video-LLMs under both Base mode and controlled evidence perturbations, including invariant-label inductions and temporal interventions on order-sensitive items.
Across 14 models, factuality remains challenging for procedural judgment and physical reasoning, whereas faithfulness errors concentrate in motion- and structure-centric skills.
Models also exhibit limited stability under invariant-label perturbations and pronounced temporal inertia under temporal intervention.

\section*{Limitations}

The induced modes use controlled operators as proxies for deployment perturbations and do not cover the full space of corruptions or unreliable conditioning.
Temporal Intervention is tested on an order-sensitive subset using shuffling and reversal; it probes temporal reliance but does not localize the cues driving model decisions.

\bibliography{custom}

\appendix

\newpage
\section{Appendix}
\renewcommand{\thetable}{A\arabic{table}}
\renewcommand{\thefigure}{A\arabic{figure}}
\setcounter{table}{0}
\setcounter{figure}{0}
\label{sec:appendix}


\begin{table*}[tbh]
    \centering
    \small
    \renewcommand{\arraystretch}{1.08}
    \setlength{\tabcolsep}{6pt}
    
    \resizebox{\textwidth}{!}{%
        \begin{tabular}{>{\centering\arraybackslash}p{4.4cm} >{\centering\arraybackslash}p{7.6cm}}
        \Xhline{1.2pt}
        \textbf{Open-Source Video-LLMs} & \textbf{HF Checkpoint} \\
        \Xhline{0.8pt}
        VideoLLaMA2-7B & DAMO-NLP-SG/VideoLLaMA2-7B \\
        Qwen2.5VL-7B & Qwen/Qwen2.5-VL-7B-Instruct \\
        Qwen3VL-8B & Qwen/Qwen3-VL-8B-Instruct \\
        ShareGPT4Video-8B & Lin-Chen/sharegpt4video-8b \\
        NVILA-8B & Efficient-Large-Model/NVILA-8B \\
        InternVL3-8B & OpenGVLab/InternVL3-8B \\
        InternVL3.5-8B & OpenGVLab/InternVL3\_5-8B \\
        PLLaVA-13B & ermu2001/pllava-13b \\
        Qwen2.5VL-32B & Qwen/Qwen2.5-VL-32B-Instruct \\
        Qwen3VL-32B & Qwen/Qwen3-VL-32B-Instruct \\
        PLLaVA-34B & ermu2001/pllava-34b \\
        Tarsier-34B & omni-research/Tarsier-34b \\
        \Xhline{1.2pt}
        \end{tabular}%
    }
    \caption{Simplified configurations for open-source multimodal foundation models used in the evaluation.}
    \label{tab:os_models_appendix_simple}
\end{table*}

\subsection{Detailed Taxonomy and Examples}

In this appendix, we provide the comprehensive definitions and detailed examples for our proposed hallucination taxonomy. Table~\ref{tab:faithful_taxonomy} details the three-level hierarchy for \textbf{Faithfulness Hallucination}. Table~\ref{tab:factual_taxonomy} presents the categorization for \textbf{Factuality Hallucination}.

\subsection{TSS: Order-Sensitive vs.\ Non-Order-Sensitive}
\label{sec:tss_order}

To validate that TSS provides meaningful separation, Table~\ref{tab:tss_order} compares TSS computed on the order-sensitive subset ($\mathcal{S}_{\text{order}}$) and non-order-sensitive items ($\mathcal{S}_{\text{nonorder}}$) for models with non-trivial temporal responsiveness. For all models, TSS on $\mathcal{S}_{\text{order}}$ substantially exceeds TSS on $\mathcal{S}_{\text{nonorder}}$, confirming that temporal intervention is more impactful on genuinely time-dependent items and that the order-sensitivity annotation provides a useful signal.

\begin{table}[tbh]
\centering
\small
\setlength{\tabcolsep}{5pt}
\renewcommand{\arraystretch}{1.05}
\resizebox{\linewidth}{!}{%
\begin{tabular}{lcc}
\toprule
\textbf{Model} & \textbf{TSS} ($\mathcal{S}_{\text{order}}$) & \textbf{TSS} ($\mathcal{S}_{\text{nonorder}}$) \\
\midrule
Qwen2.5-VL-7B & 20.5\% & 4.5\% \\
Qwen2.5-VL-32B & 19.5\% & 6.0\% \\
Qwen3-VL-8B & 28.0\% & 9.0\% \\
Qwen3-VL-32B & 25.0\% & 8.0\% \\
InternVL3-8B & 14.0\% & 6.0\% \\
InternVL3.5-8B & 22.5\% & 5.5\% \\
Tarsier-34B & 21.5\% & 7.5\% \\
\bottomrule
\end{tabular}%
}
\caption{\textbf{TSS on order-sensitive vs.\ non-order-sensitive items.} Models with non-trivial temporal responsiveness show meaningful separation between subsets.}
\label{tab:tss_order}
\end{table}

\subsection{Human Validity Audit for Invariant-Label Operators}
\label{sec:human_validity}

A key assumption behind Modes~II--III is that the perturbation operators are \emph{label-preserving}, i.e., although the input is degraded or corrupted, the correct answer should remain unchanged. To verify that this assumption is aligned with human judgment, we conduct a human validity audit over all invariant-label operators.

For each operator, we randomly sample 200 perturbed video--QA pairs and ask three independent annotators, blinded to the original gold labels, to first judge whether the perturbed instance remains answerable and, if so, to select the correct multiple-choice option. We aggregate annotations by majority vote and report two measurements: \emph{answerability}, the fraction of perturbed instances that remain well-posed, and \emph{label preservation}, the fraction of answerable instances whose majority-vote label matches the original gold answer. This protocol provides a direct human check of the invariant-label assumption used by RR.

Table~\ref{tab:operator_validity} shows that the single-operator perturbations used in Modes~II--III are largely consistent with human judgments: answerability ranges from 98.5\% to 100.0\%, and label preservation ranges from 99.5\% to 100.0\%. These results support the use of RR as a reliability metric under visual degradation and evidence corruption.

Caption injection is additionally analyzed at the subtitle-subtype level, because it mixes two distinct forms of textual interference: content-irrelevant subtitles and intentionally misleading subtitles. As shown in Table~\ref{tab:caption_validity}, irrelevant captions fully preserve answerability and labels, whereas misleading captions yield slightly lower answerability (94.0\%) and label preservation (94.5\%), which is expected given their adversarial design. Even so, the rates remain sufficiently high to justify treating caption injection as an approximately label-preserving operator in our evaluation.

\begin{table}[tbh]
\centering
\small
\setlength{\tabcolsep}{4pt}
\renewcommand{\arraystretch}{1.05}
\begin{tabular*}{\linewidth}{@{}l@{\extracolsep{\fill}}cc@{}}
\toprule
\textbf{Operator} & \textbf{Answerability} & \textbf{Label-pres.} \\
\midrule
Compression (cmp) & 99.5\% & 100.0\% \\
Subtitle corruption (sub) & 99.0\% & 99.5\% \\
Gaussian noise (gau) & 100.0\% & 100.0\% \\
Motion blur (mb) & 100.0\% & 99.5\% \\
Adversarial noise (adv) & 98.5\% & 99.5\% \\
\bottomrule
\end{tabular*}
\caption{\textbf{Human validity audit for invariant-label operators.} Each operator is audited on 200 perturbed video--QA pairs with three blinded annotators and majority-vote aggregation. High answerability and label-preservation rates indicate strong agreement between the automatic invariant-label evaluation and human judgments.}
\label{tab:operator_validity}
\end{table}

\begin{table}[tbh]
\centering
\small
\setlength{\tabcolsep}{4pt}
\renewcommand{\arraystretch}{1.05}
\begin{tabular*}{\linewidth}{@{}l@{\extracolsep{\fill}}cc@{}}
\toprule
\textbf{Subtype} & \textbf{Answerability} & \textbf{Label-pres.} \\
\midrule
Irrelevant & 100.0\% & 100.0\% \\
Misleading & 94.0\% & 94.5\% \\
\bottomrule
\end{tabular*}
\caption{\textbf{Caption injection validity by subtype.} Irrelevant captions fully preserve answerability and labels. Misleading captions show slightly lower rates, consistent with their adversarial design.}
\label{tab:caption_validity}
\end{table}

\subsection{Factuality vs. Knowledge Gaps}
\label{sec:factuality_diagnostic}

To probe whether factuality errors are predominantly hallucination-like (high-confidence wrong) or knowledge-gap-like (low-confidence/uncertain wrong), we conduct a behavioral diagnostic on 369 factuality questions across 8 open-source models, yielding 2,952 model-question pairs (8×369).
Each model is evaluated with standard prompting (clean run) and additionally asked to report a self-assessed confidence score (diagnostic run).
Wrong instances are classified as: \emph{hallucination-like} if the reported confidence is $\geq$70\%, \emph{knowledge-gap-like} if the confidence is $\leq$40\% or the model expresses explicit uncertainty, and \emph{other} otherwise.

Among 888 analyzable wrong instances, 797 (89.75\%) exhibit hallucination-like behavior, while only 39 (4.39\%) show knowledge-gap-like patterns.
This suggests that factuality failures in current Video-LLMs are overwhelmingly overconfident, producing wrong answers with high self-reported certainty rather than acknowledging uncertainty.
We note that this is a behavioral diagnostic rather than a causal attribution---knowledge gaps may also manifest as overconfident guessing.

\subsection{Implementation Details for Operators}
\label{sec:deployment_impl}

\paragraph{Compression (cmp).}
Videos are re-encoded using FFmpeg with a target bitrate retaining approximately 15.19\% of the original (reducing storage from 2.83\,GB to 0.43\,GB on average), simulating the lossy compression commonly applied by video-sharing platforms.

\paragraph{Subtitle Corruption (sub).}
Noisy ASR-like subtitles are generated by injecting character-level errors (substitution, deletion, insertion) into ground-truth transcriptions, with intentional subtitle--video desynchronization (random temporal shifts of 0.5--2.0 seconds). This simulates the unreliable OCR/ASR outputs and timing misalignment encountered in deployment.

\paragraph{Gaussian Noise (gau).}Gaussian noise is applied uniformly at the frame level to simulate sensor noise or low-light recording conditions. We add zero-mean Gaussian noise to the RGB channels of each frame. The perturbation variance is controlled to keep the noise bounded, ensuring that the high-level semantic content required to answer the question is preserved while degrading low-level perceptual clarity.

\paragraph{Motion Blur (mb).}Motion blur is applied uniformly across all frames to simulate camera shake or fast-moving subjects. We apply a linear motion blur filter with randomized kernel sizes and angles to the spatial dimensions of each frame. This operation reduces visual sharpness and introduces perceptual ambiguity without altering the underlying dynamic events.

\paragraph{Caption Injection (cap).}To create cross-modal conflicts, we overlay synthesized subtitles onto randomly sampled video segments. As analyzed in Appendix~\ref{sec:human_validity}, the injected subtitles consist of a 4:1 mixture of content-irrelevant sentences and LLM-generated misleading statements. The misleading statements are explicitly conditioned on the question and the ground-truth answer option to form a plausible but incorrect textual cue (e.g., pairing a video of opening a door'' with the text closing the door''), testing the model's ability to prioritize authentic visual evidence over misleading text.

\paragraph{Adversarial Noise (adv).}We apply a transfer-based black-box perturbation targeting the visual encoder, generated using the MI-FGSM~\cite{Dong_2018_CVPR_MIFGSM} algorithm. To ensure the adversarial noise generalizes across different Video-LLMs architectures rather than overfitting to a single proxy, the perturbations are crafted using a proxy ensemble of varied visual encoders (InternVL3-8B~\cite{zhu2025internvl3}, Qwen3VL-8B~\cite{Qwen3-VL}, and Video-MAE~\cite{tong2022videomae}).

\paragraph{Temporal Intervention (shu \& rev).}Applied exclusively to the order-sensitive subset ($\mathcal{S}_{\text{order}}$), these operators disrupt the temporal structure essential for correctness. \textbf{Shuffling (shu)} randomly permutes the sequence of frames across the video, disrupting both local motion and global event order. \textbf{Reversal (rev)} strictly inverts the chronological order of the frames from end to start, reversing the direction of actions and state transitions. Both operators render the original ground-truth label invalid by destroying the necessary causal dependencies.

\begin{table*}[tp]
    \centering
    \renewcommand{\arraystretch}{1.35} 
    \setlength{\tabcolsep}{8pt}
    
    \resizebox{1.0\textwidth}{!}{%
        \begin{tabular}{p{7cm}|p{10cm}} 
        \Xhline{1.2pt}
        \textbf{Task \& Definition} & \textbf{Example} \\
        \Xhline{0.8pt}

        \multicolumn{2}{l}{\cellcolor{gray!15}\textbf{1. Static Entities \& Attributes }} \\ 
        \hline

        \textbf{1.1 Entity Recognition} \newline 
        \small{\textit{Identify an entity label supported by video evidence.}}
        & \textit{\darkblue{Which human organ is visible in the video?}} \newline 
          (A) Stomach \quad (B) Liver \quad (C) Lungs \quad (D) Kidneys \\ 
        \hline

        \textbf{1.2 Unique Entity Counting} \newline 
        \small{\textit{Count the number of distinct entity instances.}}
        & \textit{\darkblue{How many babies does the lion mother have in the video?}} \newline 
          (A) 4 \quad (B) 3 \quad (C) 2 \quad (D) 5 \\ 
        \hline

        \textbf{1.3 Temporal Attributes Recognition} \newline 
        \small{\textit{Attribute query with an explicit temporal/event anchor.}}
        & \textit{\darkblue{What is the color of the second object that enters the scene?}} \newline 
          (A) Brown \quad (B) Cyan \quad (C) Gray \quad (D) Purple \\ 
        \hline

        \textbf{1.4 Static Attributes Recognition} \newline 
        \small{\textit{Attribute query without explicit temporal/event anchor.}}
        & \textit{\darkblue{What color is the T-shirt worn by the boy?}} \newline 
          (A) Yellow \quad (B) White \quad (C) Black \quad (D) Blue \\ 
        \hline

        \textbf{1.5 Scene Text Recognition} \newline 
        \small{\textit{Detecting and reading text that appears in the video.}}
        & \textit{\darkblue{What is written on the first made keychain?}} \newline 
          (A) Google (B) YouTube (C) Facebook (D) Spotify \\ 
        \hline

        \multicolumn{2}{l}{\cellcolor{gray!15}\textbf{2. Dynamic Actions \& Motions}} \\ 
        \hline

        \textbf{2.1 Action Recognition} \newline 
        \small{\textit{Discriminate fine-grained / near-confusable action.}}
        & \textit{\darkblue{Which description correctly matches the actions?}} \newline 
          (A) Attacking \quad (B) Chasing \quad (C) Running (D) Stealing \\ 
        \hline

        \textbf{2.2 Repetitive Action Counting} \newline 
        \small{\textit{Count occurrences of actions across the video.}}
        & \textit{\darkblue{How many times does the hammer's handle hit the floor throughout the video?}} \newline 
          (A) 4 (B) 3 (C) 1 (D) 0 (E) 5 (F) 2\\ 
        \hline

        \textbf{2.3 Motion Attributes Recognition} \newline 
        \small{\textit{Identify motion attributes (direction, rotation,speed, trajectory) of an entity.}}
        & \textit{\darkblue{In what direction(s) is the wheel rotating?}} \newline 
          (A) Counter-clockwise throughout (B) Clockwise throughout (C) No rotation (D) Counter-clockwise then clockwise. \\ 
        \hline

        \multicolumn{2}{l}{\cellcolor{gray!15}\textbf{3. Spatio-Temporal Relations}} \\ 
        \hline

        \textbf{3.1 Spatial Relation Recognition} \newline 
        \small{\textit{Infer spatial relations or spatial locations in the video.}}
        & \textit{\darkblue{Where is the hidden object at the end of the game?}} \newline 
          (A) Under 1st left \quad (B) Under 3rd left (C) Under 2nd from left\\ 
        \hline

        \textbf{3.2 Temporal Relation Recognition} \newline 
        \small{\textit{Determine the temporal relations among events.}}
        & \textit{\darkblue{What happened after the person sat on the sofa/couch?}} \newline 
         (A) Opened the box (B) Opened the door (C) Sat on the floor (D) Put down the clothes. \\ 
         \hline

        \textbf{3.3 State Transition Detection} \newline 
        \small{\textit{Determine whether a state variable changes over time.}}
        & \textit{\darkblue{Is the bag empty at the end?}} \newline 
          (A) Yes \quad (B) No \quad (C) The person doesn't interact with a bag \\ 
        \hline

        \textbf{3.4 Temporal Localization} \newline 
        \small{\textit{Localize when an action/state/event holds within the video.}}
        & \textit{\darkblue{When in the video does the action occur?}} \newline 
          (A) Beginning \quad (B) Middle \quad (C) End \quad (D) Throughout \\ 
        \hline

        \Xhline{1.2pt}
        \end{tabular}%
    }
    \vspace{-0.2cm}

    \caption{Faithfulness Hallucination Taxonomy.}
    \label{tab:faithful_taxonomy}
    
\end{table*}

\begin{table*}[tp]
    \centering
    \renewcommand{\arraystretch}{1.2} 
    \setlength{\tabcolsep}{6pt}
    
    \resizebox{1.0\textwidth}{!}{%
        \begin{tabular}{p{6.5cm}|p{10.5cm}} 
        \Xhline{1.2pt}
        \textbf{Category \& Definition} & \textbf{Example} \\
        \Xhline{0.8pt}

        \multicolumn{2}{l}{\cellcolor{gray!15}\textbf{1. Domain Knowledge (Know-WHAT)}} \\ 
        \hline

        \textbf{1.1 Cultural Event Recognition} \newline 
        \small{\textit{Knowledge of traditions, festivals, artistic heritage, and customs.}}
        & \textit{\darkblue{When is the festival in the video usually celebrated?}} \newline 
          (A) 9th day of 9th lunar month  (B) 8th day of 12th lunar month  (C)5th day of 5th lunar month  (D) 1st day of 1st lunar month \\ 
        \hline

        \textbf{1.2 Historical Background Identification} \newline 
        \small{\textit{Knowledge of past events, chronologies, archaeology, and historical figures.}}
        & \textit{\darkblue{When was the Site shown in the video discovered?}} \newline 
          (A) 1928-1929  (B) 1925-1926  (C) 1920-1922  (D) 1934-1938 \\ 
        \hline

        \textbf{1.3 Geospatial Localization} \newline 
        \small{\textit{Knowledge of geospatial locations, landmarks, and regional characteristics.}}
        & \textit{\darkblue{Which city is this video most likely recorded in? }} \newline 
          (A)  San Diego  (B) Phoenix  (C) Tucson  (D) Las Vegas \\ 
        \hline

        \textbf{1.4 Entertainment-Related Recognition} \newline 
        \small{\textit{Knowledge of movies, TV, music, celebrities, and sports.}}
        & \textit{\darkblue{What genre of movie is portrayed in the video?}} \newline 
          (A) Reality TV (B) Documentary  (C) Thriller  (D) Romance \\ 
        \hline

        \multicolumn{2}{l}{\cellcolor{gray!15}\textbf{2. Procedural Knowledge (Know-HOW)}} \\ 
        \hline

        \textbf{2.1 Electronic Procedural Judgement} \newline 
        \small{\textit{Logical steps for repairing computers and electronic hardware.}}
        & \textit{\darkblue{Is the sequence of steps about task <Replace Laptop Screen> shown logical?}} \newline 
          (A) Yes, the sequence is correct \newline 
          (B) No, the sequence is incorrect \\ 
        \hline
        
        \textbf{2.2 Mechanical Procedural Judgement} \newline 
        \small{\textit{Sequence and tools for vehicle maintenance and mechanical repair.}}
        & \textit{\darkblue{Is the sequence of steps about task <Replace Car Door Handle> shown logical?}} \newline 
          (A) Yes, the sequence is correct \newline 
          (B) No, the sequence is incorrect \\ 
        \hline

        \textbf{2.3 Domestic Procedural Judgement} \newline 
        \small{\textit{Workflow for household DIY, plumbing, and furniture repair.}}
        & \textit{\darkblue{Is the sequence of steps about task <Replace Light Socket> shown logical?}} \newline 
          (A) Yes, the sequence is correct \newline 
          (B) No, the sequence is incorrect \\ 
        \hline

        \textbf{2.4 Clinical Procedural Judgement} \newline 
        \small{\textit{Medical protocols for examinations, first aid, and clinical operations.}}
        & \textit{\darkblue{Is the sequence of steps about task <Measure the Jaw Opening> shown logical?}} \newline 
          (A) Yes, the sequence is correct \quad (B) No, the sequence is incorrect \\ 
        \hline
        \multicolumn{2}{l}{\cellcolor{gray!15}\textbf{3. Physical Knowledge (Know-WHY) }} \\ 
        \hline

        \textbf{3.1 Newtonian Mechanical Reasoning} \newline 
        \small{\textit{Newtonian laws: gravity, friction, collision, and momentum.}}
        & \textit{\darkblue{Analyze the physical dynamics in the video, which statement best describes the adherence to physical laws?}} \newline 
          (A) Fully Plausible  (B)Violation of Mechanical Dynamics  (C) Violation of Material Properties (D) Violation of Fluid Dynamics \\ 
        \hline

        \textbf{3.2 Fluid Mechanics Reasoning} \newline 
        \small{\textit{Behavior and interaction of liquids, smoke, and fire.}}
        & \textit{\darkblue{Analyze the physical dynamics; which statement best describes the adherence to laws?}} \newline 
          (A) Fully Plausible (B) Violation of Mechanics (C) Violation of Fluids (D) Violation of Materials \\ 
        \hline
        
        \textbf{3.3 Material Property Reasoning} \newline 
        \small{\textit{Realism of deformation, hardness, and fracture of solids.}}
        & \textit{\darkblue{Analyze the physical dynamics; which statement best describes the adherence to laws?}} \newline 
          (A) Fully Plausible (B) Violation of Mechanics (C) Violation of Fluids (D) Violation of Materials \\ 
        \hline

        \textbf{3.4 Spatio-temporal Continuity Reasoning} \newline 
        \small{\textit{Continuity and permanence of objects under occlusion or movement.}}
        & \textit{\darkblue{Analyze the physical dynamics; which statement best describes the adherence to laws?}} \newline 
          (A) Fully Plausible (B) Violation of Object Permanence (C) Violation of Continuity \\ 
        \hline

        \Xhline{1.2pt}
        \end{tabular}%
    }
    \vspace{-0.2cm}
    \caption{\textbf{Factual Hallucination Taxonomy.} The framework evaluates world knowledge across three dimensions: Domain Knowledge (Know-WHAT), Procedural Knowledge (Know-HOW), and Physical Knowledge (Know-WHY).}
    \label{tab:factual_taxonomy}
\end{table*}

\end{document}